\documentclass{article}

\usepackage{microtype}
\usepackage{graphicx}
\usepackage{subcaption}
\usepackage{booktabs} 

\usepackage{hyperref}

\usepackage{amsmath}
\usepackage{amssymb}
\usepackage{mathtools}
\usepackage{amsthm}

\usepackage[capitalize,noabbrev]{cleveref}

\usepackage{pifont}

\usepackage{tabularx}
\usepackage{array}
\newcolumntype{L}[1]{>{\raggedright\arraybackslash}p{#1}}
\newcolumntype{Y}{>{\raggedright\arraybackslash}X}

\theoremstyle{plain}

\theoremstyle{definition}

\theoremstyle{remark}

\usepackage[textsize=tiny]{todonotes}

\usepackage[preprint]{icml2026}

\icmltitlerunning{Next Embedding Prediction Makes World Models Stronger}

\begin{document}

\twocolumn[
  \icmltitle{Next Embedding Prediction Makes World Models Stronger}



  \icmlsetsymbol{equal}{*}

  \begin{icmlauthorlist}
    \icmlauthor{George Bredis}{comp}
    \icmlauthor{Nikita Balagansky}{comp}
    \icmlauthor{Daniil Gavrilov}{comp}
    \icmlauthor{Ruslan Rakhimov}{comp}
  \end{icmlauthorlist}

  \icmlaffiliation{comp}{T-Tech}
  \icmlcorrespondingauthor{George Bredis}{georgy.bredis@gmail.com}
  \icmlkeywords{model-based reinforcement learning, world models, representation learning}

  \vskip 0.3in
]



\printAffiliationsAndNotice{}  

\begin{abstract}
Capturing temporal dependencies is critical for model-based reinforcement learning (MBRL) in partially observable, high-dimensional domains. We introduce NE-Dreamer, a decoder-free MBRL agent that leverages a temporal transformer to predict next-step encoder embeddings from latent state sequences, directly optimizing temporal predictive alignment in representation space. This approach enables NE-Dreamer to learn coherent, predictive state representations without reconstruction losses or auxiliary supervision. On the DeepMind Control Suite, NE-Dreamer matches or exceeds the performance of DreamerV3 and leading decoder-free agents. On a challenging subset of DMLab tasks involving memory and spatial reasoning, NE-Dreamer achieves substantial gains. These results establish next-embedding prediction with temporal transformers as an effective, scalable framework for MBRL in complex, partially observable environments.
\end{abstract}

\section{Introduction}

\begin{figure}[t]
  \centering
  \includegraphics[width=\linewidth]{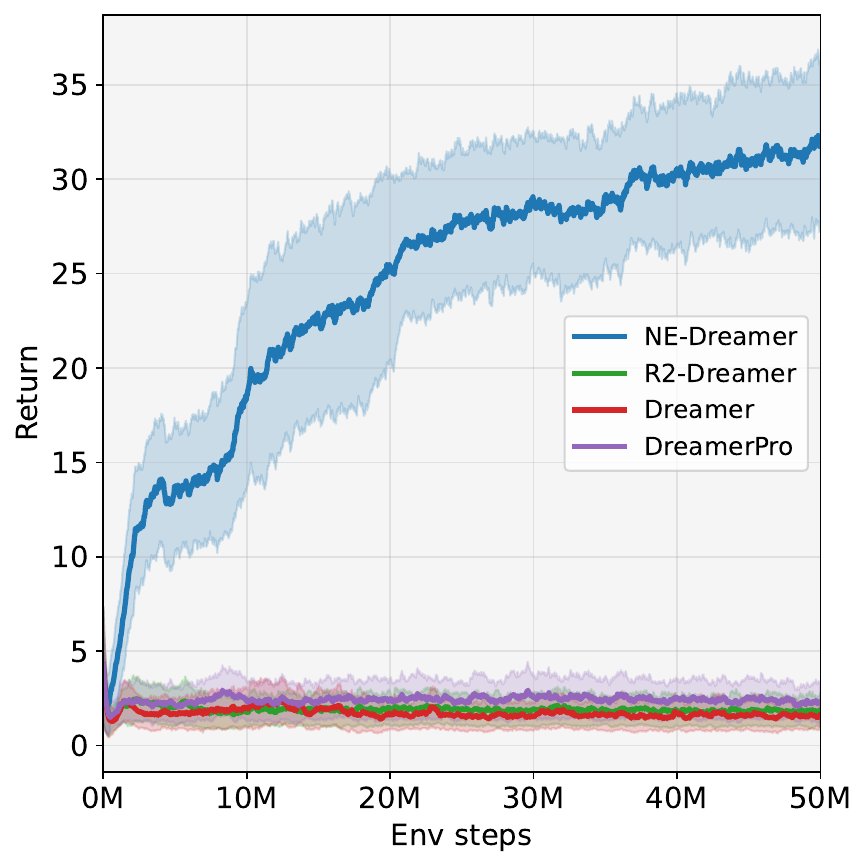}
  \caption{\textbf{DMLab Benchmark Summary.} Under matched compute and model capacity (50M environment steps; 5 seeds; 12M parameters), NE-Dreamer outperforms strong decoder-based (DreamerV3) and decoder-free world-model baselines (R2-Dreamer, DreamerPro) on the DMLab Rooms memory/navigation tasks.}
  \label{fig:teaser}
\end{figure}

Model-based reinforcement learning (MBRL) from high-dimensional observations hinges on learning a compact latent state that supports long-horizon prediction and control. This requirement becomes more important under partial observability: the agent must integrate information over time rather than react to a single frame. 

A dominant approach learns the world model with a pixel decoder, as in Dreamer, where reconstruction produces rich, control-effective features. The cost is modeling burden: reconstruction introduces a heavy generative objective, complicates optimization, and can allocate capacity to visually detailed but task-irrelevant aspects. Decoder-free methods remove the pixel decoder, training representations directly to simplify the pipeline and improve efficiency.

However, many decoder-free objectives mainly enforce \emph{instantaneous} (same-timestep) agreement. Under partial observability, instantaneous agreement is not enough: the representation must be \emph{predictive across time}. Without an explicit temporal constraint, training can drift or collapse, leading to weak long-horizon structure—failure modes that surface in memory- and navigation-heavy tasks.

\begin{figure*}[t]
  \centering
  \includegraphics[width=\linewidth]{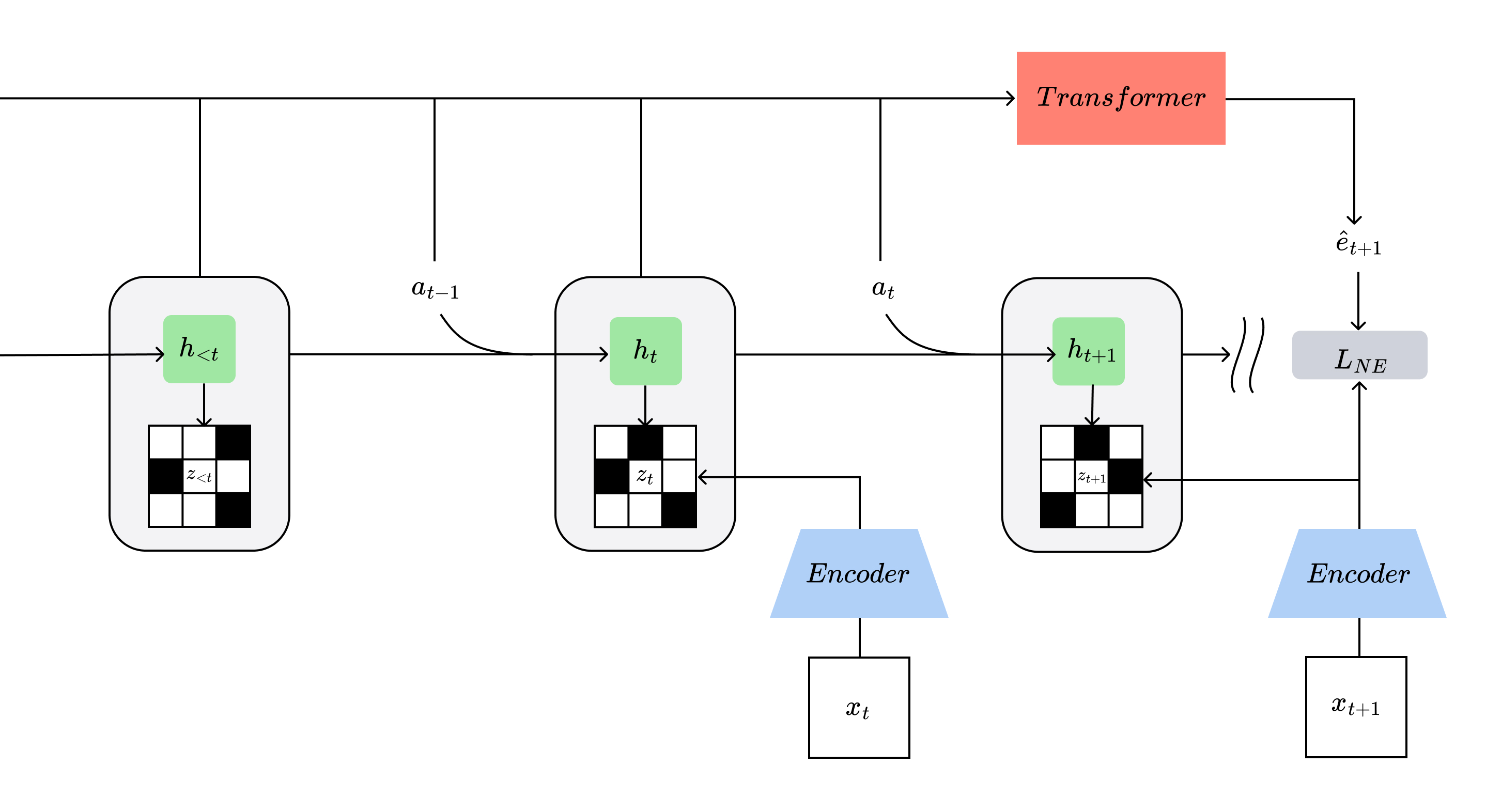}
  \caption{\textbf{Method overview.} NE-Dreamer keeps Dreamer’s RSSM dynamics and imagination-based actor--critic, but replaces same-step pixel reconstruction with \emph{next-embedding prediction} using a causal temporal transformer, improving long-horizon performance under partial observability.}
  \label{fig:method}
\end{figure*}

In this paper, we introduce NE-Dreamer, a decoder-free world model that learns by directly optimizing for \emph{temporal predictive alignment} in its latent representations. NE-Dreamer replaces pixel-level reconstruction with a simple yet powerful objective: at each timestep, a temporal transformer predicts the \emph{next} encoder embedding in the sequence, and this prediction is aligned to the actual next-step embedding using a redundancy-reduction metric (specifically, Barlow Twins in our implementation). By shifting the focus from same-timestep matching to next-step prediction, NE-Dreamer learns temporally coherent latent states without the need for pixel reconstruction, data augmentation, or auxiliary regularization. As illustrated in Figure~\ref{fig:teaser}, this design enables NE-Dreamer to achieve substantially higher performance in partially observable DMLab environments compared to prior methods of the same model size.

\medskip

Our main contributions are as follows:
\begin{enumerate}
\item We propose a decoder-free world-model objective based on \emph{next-embedding prediction}, which explicitly enforces temporal predictiveness in the learned representation.
\item We integrate a lightweight causal temporal transformer into a Dreamer-style MBRL pipeline to implement next-step prediction from history within standard RSSM training.
\item We evaluate NE-Dreamer on DeepMind Control Suite and DeepMind Lab, showing strong performance on DMC and substantial gains on memory/navigation-heavy DMLab Rooms under matched compute and model size.
\item Through targeted ablations and representation diagnostics, we isolate the gains to predictive sequence modeling (causal transformer + next-step target shift) rather than reconstruction or auxiliary tricks.
\end{enumerate}

\section{Related Work}
\label{sec:related}

\paragraph{World models for pixel control.}
Latent world models aim to learn compact states that support long-horizon prediction and decision-making from high-dimensional observations. Early work demonstrated that learning dynamics in a latent space can enable planning and control by acting ``in imagination'' from pixels~\citep{ha2018worldmodels}. PlaNet introduced the recurrent state-space model (RSSM) as a practical latent dynamics backbone for planning from images~\citep{hafner2019planet}. Building on RSSMs, the Dreamer family trains an actor--critic on imagined rollouts in latent space via \emph{latent imagination}~\citep{hafner2020dreamer,hafner2021dreamerv2,hafner2025dreamerv3}. NE-Dreamer keeps this RSSM-based control backbone and changes how the latent representation is learned.

\paragraph{Reconstruction-based world models.}
A common way to learn world-model representations is to maximize an observation likelihood (pixel reconstruction), often alongside reward and termination/continuation prediction~\citep{ha2018worldmodels,hafner2019planet,hafner2020dreamer}. Reconstruction provides dense supervision that often stabilizes optimization, but it can also allocate capacity to visually detailed factors (e.g., textures or backgrounds) that are only weakly coupled to reward. This motivates decoder-free objectives that shape the latent space directly for decision-making.

\paragraph{Decoder-free world models.}
Removing pixel reconstruction shifts the problem from modeling observations to choosing \emph{what anchors} the latent state and \emph{which time index} the learning signal targets. One family is \emph{task-oriented}: latents are optimized to support reward/value prediction and planning, with supervision induced by search or TD learning, as in MuZero and TD-MPC variants~\citep{schrittwieser2019muzero,hansen2022tdmpc,hansen2024tdmpc2}; related Dreamer-style agents also replace reconstruction with control-centric prediction objectives (e.g., MuDreamer)~\citep{burchi2024mudreamer}. A second family is \emph{representation-oriented}: models predict or align learned embeddings with self-supervised objectives, sometimes across future steps (e.g., CPC, SPR)~\citep{oord2018cpc,schwarzer2021spr,paster2021blast} and sometimes via per-timestep invariances or clustering~\citep{okada2020dreaming,okada2022dreamingv2,deng2022dreamerpro,anonymous2026r2dreamer}.

For partially observable control, even strong \emph{same-step} objectives need not make the state at time $t$ \emph{predictive} of what happens at $t{+}1$. NE-Dreamer belongs to the representation-oriented family but makes this temporal requirement explicit: a causal sequence model predicts the \emph{next} encoder embedding from history and aligns it to a stop-gradient target, turning representation learning into \emph{causal next-step prediction} rather than per-timestep agreement.

\paragraph{Representation prediction and collapse prevention.}
Predicting future embeddings is an increasingly popular alternative to reconstruction in self-supervised learning. For instance, NEPA applies next-embedding prediction with stop-gradient targets~\citep{xu2025nepa}, while I-JEPA and data2vec focus on masked prediction and context modeling~\citep{assran2023ijepa,baevski2022data2vec}. A central issue is preventing representational collapse, where the learned state becomes degenerate. In reinforcement learning, invariance via augmentations is a common stabilizer~\citep{laskin2020rad,kostrikov2020drq}, and benchmarks such as the Distracting Control Suite~\citep{stone2021dcs} make this explicit. Bootstrapping and redundancy-reduction regularizers—like those used in BYOL, SimSiam, Barlow Twins, or VICReg~\citep{grill2020byol,chen2021simsiam,zbontar2021barlow,bardes2022vicreg}—can also prevent collapse without negatives, but are usually applied to paired views at the same timestep.

NE-Dreamer generalizes these ideas to a predictive context: its causal sequence model produces a forecasted embedding $e_{t+1}$ from history, which is aligned (with, e.g., a Barlow Twins loss) to a stop-gradient target. This enforces temporal coherence in the latent space, extending redundancy reduction to future prediction rather than just within-frame invariance.

\section{Method}
\label{sec:method}

\subsection{Problem setup}
We study partially observable control from pixels. At time $t$, the environment emits an image observation $x_t$. The agent selects an action $a_t$ and receives a reward $r_t$. We also use a continuation indicator $c_t \in \{0,1\}$, where $c_t=1$ if the episode continues from $t$ to $t{+}1$ and $c_t=0$ on terminal transitions.

NE-Dreamer follows the standard Dreamer pipeline---(i) learn a latent world model from experience, and (ii) train an actor--critic on imagined rollouts in latent space---but \textbf{changes the representation objective} for the world model. Specifically, it removes pixel reconstruction and instead \textbf{predicts the next-step encoder embedding}. Using only information available up to time $t$, the model predicts $\hat e_{t+1}$ and aligns it to a stop-gradient target with a self-supervised loss (Barlow Twins in our instantiation).

\subsection{Latent world model (RSSM)}
\label{sec:rssm}
We build on a recurrent state-space model (RSSM) with a deterministic recurrent state $h_t$ and a stochastic latent $z_t$.

\begin{figure*}[th]
  \centering
  \includegraphics[width=\linewidth]{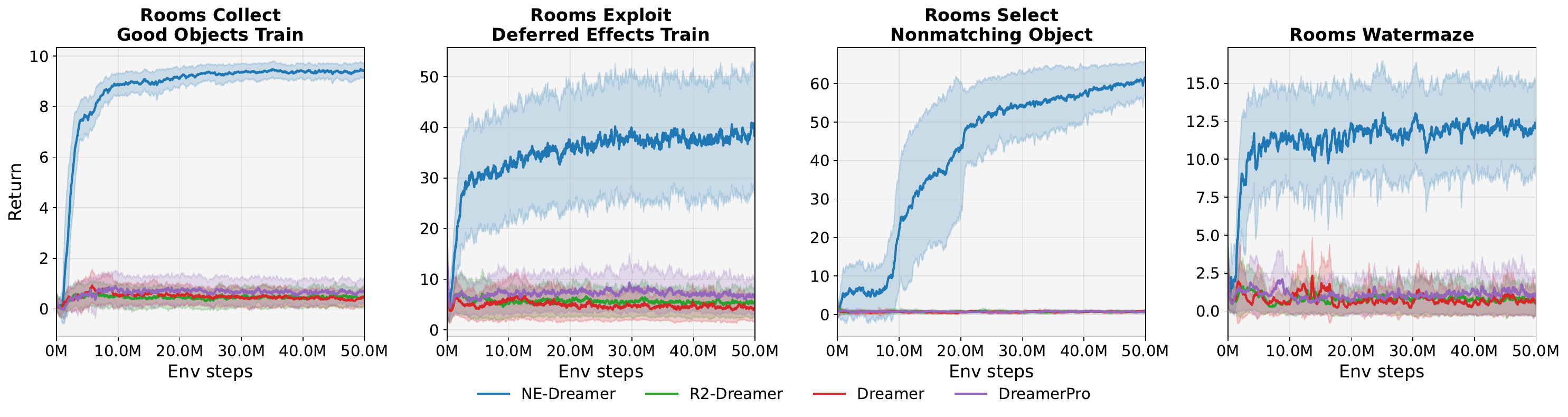}
  \caption{\textbf{DMLab Rooms: improved long-horizon memory/navigation.}
  Under matched compute and model capacity (\(50\)M environment steps; 5 seeds; 12M parameters),
  NE-Dreamer outperforms strong decoder-based (DreamerV3) and decoder-free world-model baselines (R2-Dreamer, DreamerPro) on four Rooms tasks.
  The largest gains occur when success depends on maintaining state over long horizons rather than reacting to short-lived visual cues.}
  \label{fig:dmlab}
\end{figure*}
\paragraph{Encoder and latent inference.}
An encoder maps observations to embeddings:
\begin{equation}
e_t \;=\; f_{\mathrm{enc}}(x_t).
\end{equation}
Given the previous latent state and the previous action, the RSSM updates its deterministic state:
\begin{equation}
h_t \;=\; f_{\mathrm{rec}}(h_{t-1}, z_{t-1}, a_{t-1}).
\end{equation}
It then defines a prior and posterior over the stochastic latent:
\begin{equation}
p_{\phi}(z_t \mid h_t),
\qquad
q_{\phi}(z_t \mid h_t, e_t).
\end{equation}
During world-model training we sample $z_t \sim q_{\phi}(z_t \mid h_t,e_t)$; during imagination we sample $\hat{z}_t \sim p_{\phi}(z_t \mid h_t)$.

\paragraph{Reward and continuation heads.}
As in Dreamer, the world model predicts reward and continuation:
\begin{equation}
p_{\phi}(r_t \mid h_t, z_t),
\qquad
p_{\phi}(c_t \mid h_t, z_t).
\end{equation}
Standard Dreamer also predicts observations via a pixel decoder $p_{\phi}(x_t \mid h_t, z_t)$. NE-Dreamer removes this decoder and replaces it with the next-embedding objective in Sec.~\ref{sec:ne_loss}.

\paragraph{World-model objective.}
The world model is trained with reward and continuation likelihoods, a prior--posterior regularizer, and the proposed next-embedding loss:
\begin{equation}
\mathcal{L}_{\mathrm{wm}}
\;=\;
\mathcal{L}_{\mathrm{rew}}
+
\mathcal{L}_{\mathrm{cont}}
+
\beta_{\mathrm{kl}} \, \mathcal{L}_{\mathrm{kl}}
+
\beta_{\mathrm{ne}} \, \mathcal{L}_{\mathrm{NE}}.
\end{equation}
The prediction losses are negative log-likelihoods:
\begin{equation}
\begin{aligned}
\mathcal{L}_{\mathrm{rew}}  &= -\mathbb{E}\!\left[\log p_{\phi}(r_t \mid h_t, z_t)\right], \\
\mathcal{L}_{\mathrm{cont}} &= -\mathbb{E}\!\left[\log p_{\phi}(c_t \mid h_t, z_t)\right].
\end{aligned}
\end{equation}
The KL term regularizes the posterior toward the prior:
\begin{equation}
\mathcal{L}_{\mathrm{kl}}
\;=\;
\mathbb{E}\Big[\mathrm{KL}\big(q_{\phi}(z_t \mid h_t,e_t)\;\|\;p_{\phi}(z_t \mid h_t)\big)\Big].
\end{equation}
We adopt standard Dreamer stabilizers for $\mathcal{L}_{\mathrm{kl}}$ (e.g., KL balancing / free-nats); details follow prior Dreamer practice.

\subsection{Next-embedding predictive alignment}
\label{sec:ne_loss}
NE-Dreamer trains the latent dynamics to be predictive in representation space: from history up to time $t$, it predicts the encoder embedding of the next observation and aligns the prediction to a stop-gradient target.

\paragraph{Causal next-embedding predictor.}
A causal temporal transformer $T_{\theta}$ (with a causal mask) uses only information available up to time $t$ to produce a next-step embedding prediction:
\begin{equation}
\hat{e}_{t+1} \;=\; T_{\theta}\!\big(h_{\le t}, z_{\le t}, a_{\le t}\big).
\end{equation}
The target is the next-step encoder embedding:
\begin{equation}
e^\star_{t+1} \;=\; \mathrm{sg}(e_{t+1})
\;=\; \mathrm{sg}\!\big(f_{\mathrm{enc}}(x_{t+1})\big).
\end{equation}
We write $\mathrm{sg}(\cdot)$ for stop-gradient. Gradients flow through $\hat{e}_{t+1}$ into $T_{\theta}$ and the RSSM, but not through $e^\star_{t+1}$.

\paragraph{Alignment loss (Barlow Twins).}
We instantiate $\mathcal{L}_{\mathrm{NE}}$ with a Barlow Twins redundancy-reduction objective between predicted and target embeddings. Let $\tilde{\hat{e}}_{t+1}$ and $\tilde{e}^\star_{t+1}$ denote embeddings normalized \emph{per dimension} over the set of valid transitions within each minibatch (zero mean, unit variance). Let
\begin{equation}
\mathcal{I} \doteq \{(b,t)\,:\,c_t^{(b)}=1\},
\qquad
N \doteq |\mathcal{I}|.
\end{equation}
The cross-correlation matrix is
\begin{equation}
C_{ij}
\;=\;
\frac{1}{N}\sum_{(b,t)\in\mathcal{I}}\tilde{\hat{e}}^{(b)}_{t+1,i}\,\tilde{e}^{\star(b)}_{t+1,j}.
\end{equation}
The next-embedding loss is
\begin{equation}
\mathcal{L}_{\mathrm{NE}}
\;=\;
\sum_i \big(1 - C_{ii}\big)^2
\;+\;
\lambda_{\mathrm{BT}}\sum_{i\neq j} C_{ij}^2.
\end{equation}
This objective encourages invariance (large diagonal correlations) while discouraging redundancy (small off-diagonal correlations), here applied to \emph{next-step} prediction rather than same-timestep matching.


\subsection{Actor-Critic Learning}
Like DreamerV3, NE-Dreamer learns a policy and value function in latent space by generating imagined trajectories with a world model. These imagined trajectories (of horizon $H=15$ steps) enable efficient batch actor-critic updates. We denote the imagined full latent state as $s_t = (h_t, \hat{z}_t)$, where $\hat{z}_t \sim p_{\phi}(z_t \mid h_t)$. At each imagination step, actions are sampled from the policy $\pi_\theta$ and their values are estimated by the critic $V_\psi$:
\begin{equation}
a_t \sim \pi_\theta(a_t \mid s_t), \quad V_\psi(s_t) \approx \mathbb{E}_{p_\phi, \pi_\theta}[R_t^\lambda]
\end{equation}

\textbf{Critic:} The critic predicts the distribution of $\lambda$-returns based on imagined rewards:
\begin{equation}
R_t^\lambda = r_t + \gamma c_t \big( (1-\lambda) V_\psi(s_{t+1}) + \lambda R_{t+1}^\lambda \big)
\end{equation}
\begin{equation}
\mathcal{L}_{\text{critic}}(\psi) = -\mathbb{E}_{p_\phi, \pi_\theta} \left[ \sum_{t=1}^{H} \log p_\psi(R_t^\lambda \mid s_t) \right]
\end{equation}

\textbf{Actor:} The actor maximizes normalized advantages, with $S$ as an EMA-based scale:
\begin{equation}
\begin{aligned}
\mathcal{L}_{\text{actor}}(\theta) =\;
  & -\mathbb{E}_{p_\phi,\pi_\theta} \bigg\{
      \sum_{t=1}^{H}
        \mathrm{sg}\left(
          \frac{R_t^\lambda - V_\psi(s_t)}{\max(1, S)}
        \right)
        \log \pi_\theta(a_t \mid s_t) \\
   & \qquad\qquad
        + \eta\, \mathcal{H}[\pi_\theta(a_t \mid s_t)]
     \bigg\}
\end{aligned}
\end{equation}

Here, $\text{sg}(\cdot)$ denotes the stop-gradient operator and $\eta$ is the entropy regularization coefficient.

Policy gradients are backpropagated through the world model for continuous actions. The learning procedure and all hyperparameters match DreamerV3, ensuring that observed gains stem from the representation learning objective.

\section{Experiments}
\label{sec:experiments}

\begin{figure*}[th]
  \centering
  \includegraphics[width=\linewidth]{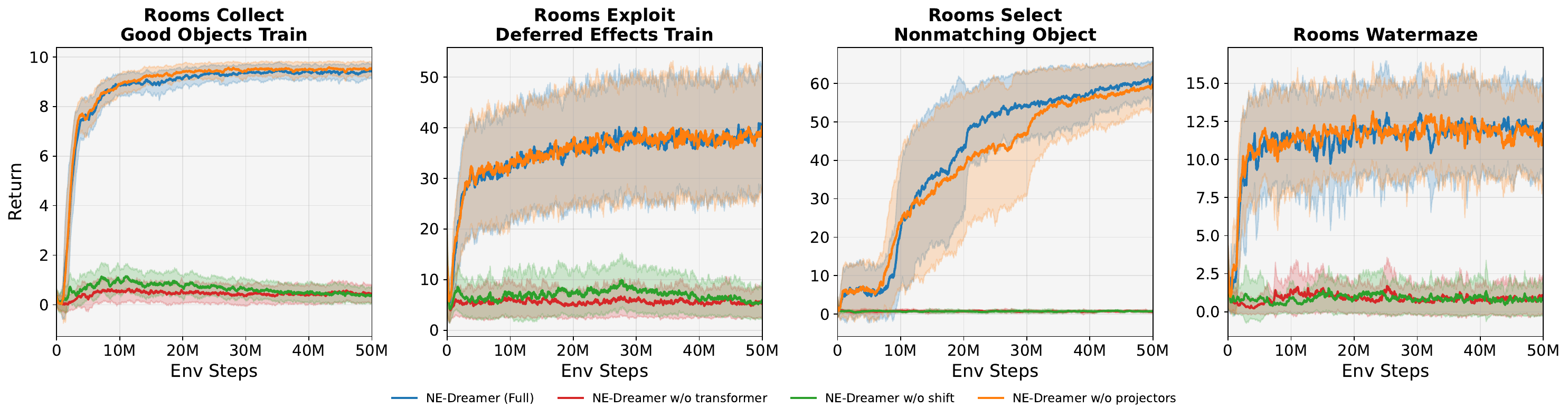}
  \caption{\textbf{Mechanism on DMLab Rooms: predictive sequence modeling is the key.}
  Under matched compute and model capacity (\(50\)M environment steps; \(5\) seeds; mean$\pm$std),
  removing the causal temporal transformer (\emph{w/o transformer}) or removing the next-step target shift (\emph{w/o shift})
  substantially reduces performance.
  Removing the lightweight projector (\emph{w/o projector}) mainly affects optimization speed/stability, with smaller impact on final returns.}
  \label{fig:dmlab_ablations}
\end{figure*}

We evaluate whether next-embedding prediction improves long-horizon control under partial observability. We structure the results around three claims:
\textbf{(C1)} NE-Dreamer improves memory/navigation performance on DMLab Rooms;
\textbf{(C2)} the gains come from \emph{predictive} sequence modeling (causal transformer + next-step target shift);
and \textbf{(C3)} removing reconstruction does not degrade standard continuous control on DMC.
Figure~\ref{fig:dmlab} (C1), Figure~\ref{fig:dmlab_ablations} (C2), and Figure~\ref{fig:dmc} (C3) provide the headline evidence.

\subsection{Experimental setup}
\label{sec:exp_setup}

\paragraph{Benchmarks and tasks.}
We evaluate all methods on two widely used RL benchmarks:
\begin{itemize}
    \item \textbf{DeepMind Lab (DMLab)}~\citep{beattie2016deepmind} is a suite of first-person 3D navigation tasks designed to test partial observability, long-horizon credit assignment, and memory. Our evaluation targets four challenging ``Rooms'' tasks that require agents to integrate information over time and reason about spatial layouts.
    \item \textbf{DeepMind Control Suite (DMC)}~\citep{tunyasuvunakool2020} is a standard benchmark for pixel-based continuous control in robotics-inspired environments. It is widely used to compare model-based RL methods, and recent advances have reached near-ceiling performance on many tasks.
\end{itemize}

\paragraph{Compared methods.}
We benchmark NE-Dreamer against representative state-of-the-art agents from three families:
\begin{itemize}
    \item \textit{Decoder-based world models:} \textbf{DreamerV3}: trains latent dynamics and policy using pixel-level reconstruction as the main representation objective.

\bigskip
\bigskip

    \item \textit{Decoder-free world models:}
    \begin{itemize}
        \item \textbf{R2-Dreamer} removes the pixel decoder and replaces reconstruction with a redundancy reduction loss (Barlow Twins) applied at the same timestep, enforcing agreement between encoder and latent via a lightweight projector.
        \item \textbf{DreamerPro} adopts a decoder-free design but uses strong data augmentations (random image shifts) to avoid representation collapse and enforce invariance.
        \item \textbf{Dreamer (no reconstruction)}- a special Dreamer variant that omits pixel reconstruction entirely, relying solely on reward, continuation, and KL objectives. This baseline tests the effect of removing explicit representation learning signals on the world model.
    \end{itemize}
    
    \item \textit{Model-free reference:} \textbf{DrQv2}: a strong pixel-based model-free RL agent that leverages strong data augmentation and direct policy/value learning from observations, providing a competitive non-model-based baseline.
\end{itemize}

All agents, including NE-Dreamer, baselines, and DrQv2 (which uses its official implementation), are evaluated under identical conditions: world-model methods share a unified PyTorch R2-Dreamer codebase with matched capacity (~12M parameters, Dreamer-S architecture), while all agents undergo the same training protocol (50M environment steps on DMLab, 1M on DMC) across five random seeds. Results are reported as mean ± standard deviation; full architectural, hyperparameter, and reproducibility details appear in Appendix~\ref{app:hypers}.

\subsection{DMLab Rooms: long-horizon memory and navigation (C1)}
\label{sec:exp_dmlab}

The DMLab Rooms benchmark directly targets the core challenge for model-based RL agents: reasoning over long temporal horizons in environments with sparse rewards and high partial observability. In these tasks, agents must integrate information across time, remember key scene elements, and plan multi-step behaviors—conditions under which standard per-timestep objectives often fail.

Figure~\ref{fig:dmlab} presents the per-task learning curves. Across all four tasks, NE-Dreamer delivers a dramatic improvement in returns—learning reliably and achieving substantially higher final performance than all baseline methods.

These results underscore two main strengths of NE-Dreamer:
\bigskip
\bigskip
\begin{itemize}
    \item \textbf{Superior temporal representation:} The use of next-embedding prediction with a temporal transformer enables the agent to maintain stable, predictive state representations over long horizons, a property directly reflected in its ability to solve complex spatial memory tasks.
    \item \textbf{Efficiency without extra complexity:} NE-Dreamer achieves these gains without pixel-level reconstruction, heavy data augmentation, or additional domain-specific tuning. All methods operate under identical architecture and training budgets, highlighting the effectiveness of our approach rather than differences in model capacity or optimization.
\end{itemize}


\begin{figure*}[t]
  \centering
  \includegraphics[width=\linewidth]{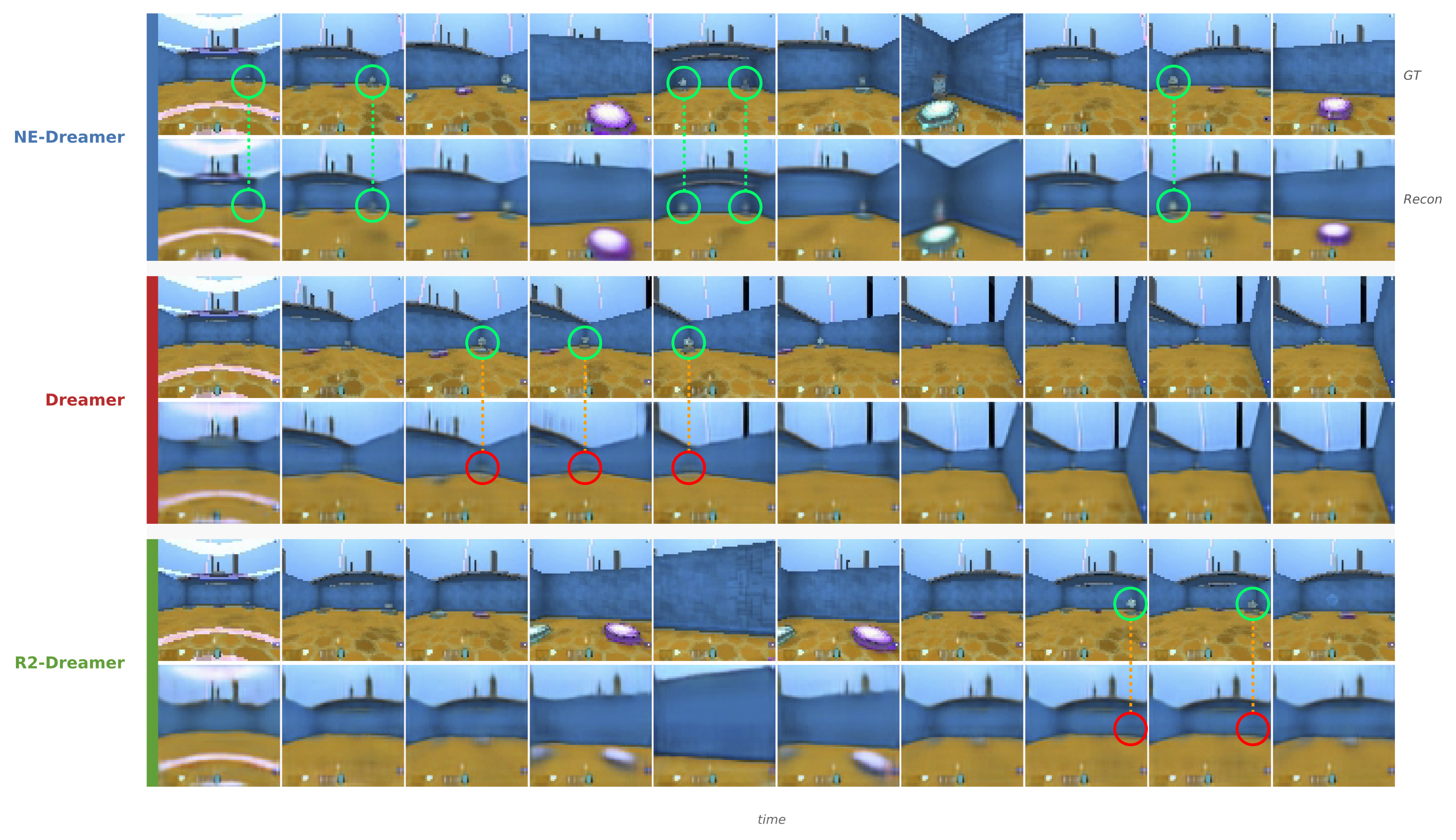}
  \caption{\textbf{Post-hoc decoder reconstruction reveals temporal consistency.}
  Rows show ground-truth observations (GT) and reconstructions from a post-hoc decoder trained on frozen latents.
  NE-Dreamer preserves task-relevant objects and spatial layout consistently over time (marked green circles),
  while same-timestep methods (Dreamer, R2-Dreamer) exhibit temporal inconsistency, where task-specific attributes
  appear transiently and then fade (marked red circles).}
  \label{fig:repr_recon}
\end{figure*}

\subsection{DMLab Rooms ablations: isolating the mechanism (C2)}
\label{sec:exp_ablation}

To isolate the key contributors to NE-Dreamer's performance, we systematically ablate three architectural and objective choices, keeping the rest of the pipeline strictly unchanged. The results, shown in Figure~\ref{fig:dmlab_ablations}, highlight the critical importance of both the temporal transformer and the next-step prediction target.

\textbf{No transformer:}  
When the temporal transformer is removed, the model regresses to using a simple feedforward or shallow architecture for sequence modeling. As shown by the red curve, performance collapses on all tasks, highlighting that causal sequence modeling is indispensable for partially observable environments. The agent fails to maintain useful temporal state, suggesting that the transformer’s sequence modeling capacity is important in this regime.

\textbf{No next-step shift:}  
Here, the model is trained to match the current-step embedding (as in most bootstrapped or instantaneous self-supervised objectives), rather than to predict the next-step target, while maintaining temporal transformer. This ablation demonstrates a nearly complete loss of the gains seen in the full method. The result points directly to the need for temporal prediction—not merely matching or reconstructing current observations, but explicitly encouraging the model to anticipate future latent structure.

\textbf{No projector:}  
The lightweight projection head before transformer is removed in this setting. Such a change leads to only a minor reduction in asymptotic performance. This suggests that while the projector may aid optimization, by smoothing the alignment objective or improving conditioning, it is not fundamentally responsible for the observed gains.

Together, these ablations show that NE-Dreamer's core mechanism is the combination of a causal temporal transformer and a next-step prediction objective. The model’s success is not due to auxiliary tricks or architectural tweaks, but to its direct enforcement of temporal predictive alignment over latent trajectories.

\subsection{DMC: no regression without reconstruction (C3)}
\label{sec:exp_dmc}

\begin{figure*}[t]
  \centering
  \includegraphics[width=\linewidth]{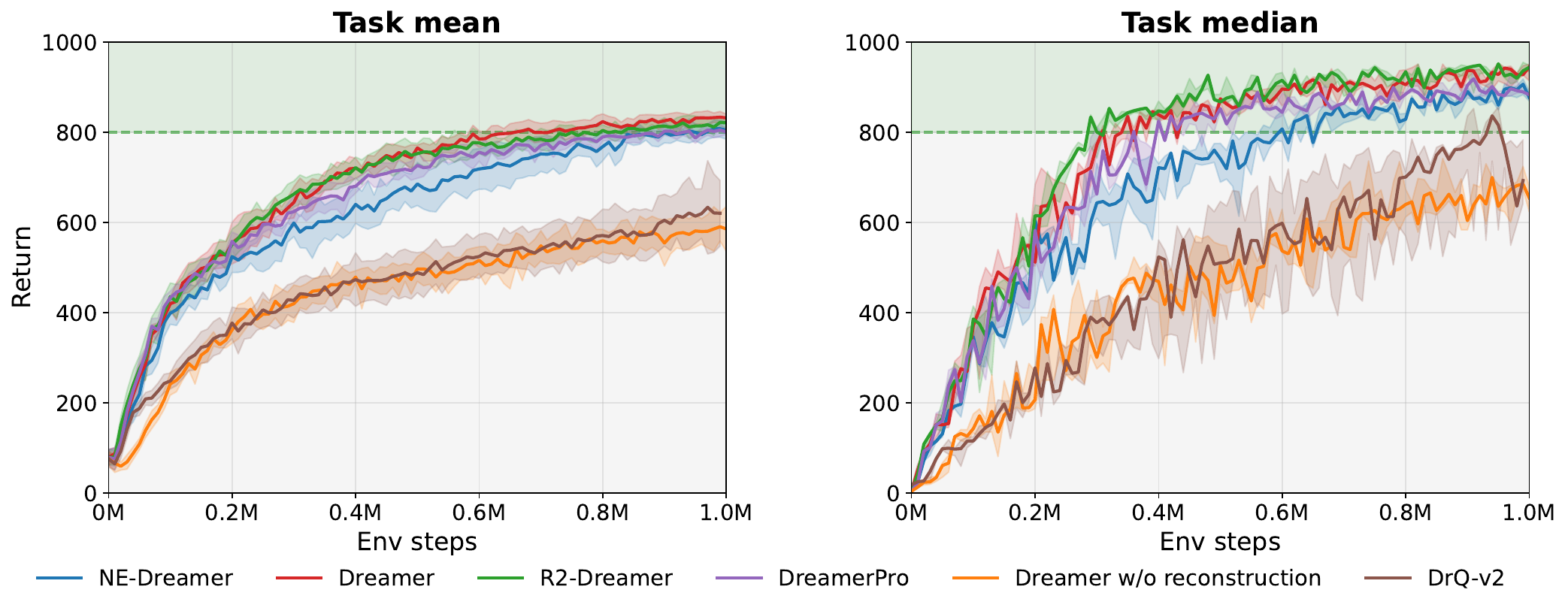}
  \caption{\textbf{DMC: removing reconstruction does not hurt standard control.}
  On near-saturated pixel-based continuous-control benchmarks, NE-Dreamer matches or slightly exceeds strong decoder-based (DreamerV3) and decoder-free world-model baselines (R2-Dreamer, DreamerPro) under a unified protocol (1M environment steps; 5 seeds; 12M parameters). Per-task learning curves can be found in Appendix~\ref{app:dmc} }
  \label{fig:dmc}
\end{figure*}

We include DMC as a calibration point. Under the unified protocol, NE-Dreamer matches DreamerV3 and competitive decoder-free baselines (Figure~\ref{fig:dmc}),
supporting the practical takeaway that replacing reconstruction with next-embedding prediction improves the hard regime (DMLab) \emph{without} sacrificing standard continuous-control performance.

\subsection{Representation diagnostics}
\label{sec:exp_repr}

To interpret what information is encoded in the learned latent state, we perform a lightweight diagnostic: we train a post-hoc pixel decoder to reconstruct observations from frozen latent representations. Importantly, this decoder is \emph{not} used during agent training and serves only as an analysis tool.

As shown in Figure~\ref{fig:repr_recon}, NE-Dreamer's latent representations enable reconstructions that preserve object identity, spatial layout, and task-relevant features consistently across time. In contrast, decoder-based Dreamer and decoder-free R2-Dreamer exhibit a characteristic failure mode: task-specific attributes (e.g., the relevant object in a room) may be present in one timestep but disappear or degrade in subsequent latents, even when the underlying scene has not changed.

NE-Dreamer’s next-embedding prediction objective enforces temporal stability by training the world model to predict the \emph{next encoder embedding} from history, which encourages the latent state to retain information that is predictive of what comes next. In contrast, same-timestep reconstruction or alignment objectives can allow latent drift toward transient visual details. Consequently, NE-Dreamer learns representations that prioritize persistent, decision-relevant structure, making it better suited for memory, planning, and long-horizon control.

\section{Discussion}

NE-Dreamer abandons pixel reconstruction in favor of direct next-embedding prediction: the model learns to predict the next encoder embedding $\hat e_{t+1}$ from history and aligns it to a stop-gradient target $e^\star_{t+1}$. We use the Barlow Twins (BT) objective to ensure stability and avoid collapse, but any alignment loss that encourages both expressiveness and non-degenerate solutions could be substituted.

A causal temporal transformer critically enables world models to compress history into only those latent features predictive of future states—yielding robustness to partial observability. Its architecture inherently supports multi-step prediction (latent overshooting), allowing efficient training of long-horizon dependencies without additional rollout cost.

NE-Dreamer delivers consistent, substantial gains on memory- and planning-intensive DMLab Rooms tasks—outperforming both decoder-free and strong decoder-based baselines at equal model size and compute. These improvements arise from temporal predictive alignment with a sequence model, not larger architectures or aggressive tuning. On standard DMC benchmarks, NE-Dreamer matches prior methods, confirming that its advantages in harder domains incur no regression elsewhere.

One limitation is that our experiments focus on environments where long-term structure, rather than fine visual detail, is the primary challenge. Whether decoder-free, prediction-based objectives can match reconstruction in high-fidelity tasks remains open. Future work should explore alternative alignment losses and test NE-Dreamer in visually complex domains.

Overall, our results establish next-embedding prediction with a causal transformer as a practical, scalable foundation for robust representation learning in model-based RL.

\section{Conclusion}

We presented NE-Dreamer, a decoder-free Dreamer-style agent that learns world-model representations by predicting and aligning the \emph{next} encoder embedding using a causal temporal transformer. NE-Dreamer improves long-horizon memory/navigation in DeepMind Lab Rooms while matching strong baselines on the DeepMind Control Suite, and ablations attribute these gains to predictive sequence modeling (causal transformer and next-step target shift), not reconstruction.




\clearpage

\bibliography{main}
\bibliographystyle{icml2026}

\newpage
\appendix
\section{Technical details}
Table~\ref{app:hypers} summarizes the primary hyperparameters used in this study. These settings are
primarily based on those of DreamerV3, with minimal modifications related to the proposed
representation learning objective.
\label{app:hypers}
\begin{table*}[h]
\caption{Main hyperparameters. Our settings are identical to DreamerV3 unless otherwise noted. All method-based hyperparameters identical to original implementations too.}
\scriptsize
\setlength{\tabcolsep}{20pt}
\renewcommand{\arraystretch}{1.25}
\centering
\hfill \break
\begin{tabular}{lcc}
\toprule
Parameter & Symbol & Setting\\ 
\midrule
General & &\\
Replay Buffer Capacity & --- & $5 \times 10^{6}$ \\
Batch Size & B & 16 \\
Batch Length & T & 64 \\
Optimizer & --- & Adam \\
Activation & --- & RMSNorm + SiLU \\
Model Size & --- & DreamerV3 Small \\ 
Input Image Resolution & --- & $64 \times 64$ RGB \\
Replayed Steps per Policy Step & --- & 512 (DMC) / 32 (DMLab) \\
Environment Instances & --- & 16 (DMC) / 16 (DMLab) \\
Action Repeat & --- & 2 (DMC) / 4 (DMLab) \\
\midrule
World Model & & \\
Number of Latents & --- & 32 \\
Classes per Latent & --- & 32 \\
Prediction Loss Scale & $\beta_{pred}$ & 1.0 \\
Dynamics Loss Scale & $\beta_{dyn}$ & 1.0 \\
Representation Loss Scale & $\beta_{rep}$ & 0.1 \\
Learning Rate & $\alpha$ & $4 \times 10^{-5}$ \\
Adam Betas & $\beta_{1}$, $\beta_{2}$ & 0.9, 0.999 \\
Adam Epsilon & $\epsilon$ & $1 \times 10^{-20}$ \\
Gradient Clipping & AGC & 0.3 \\
Slow Value Momentum & --- & 0.02 \\
Model Return Lambda & $\lambda$ & 0.95 \\
\midrule
Actor Critic & & \\
Imagination Horizon & H & 15 \\
Discount & $\gamma$ & 0.85 \\
Return Lambda & $\lambda$ & 0.95 \\
Critic EMA Decay & --- & 0.98 \\
Critic EMA regularizer & --- & 1.0 \\
Return Normalization Percentiles & --- & $5^{th}$ and $95^{th}$ \\
Return Normalization Decay & --- & 0.99 \\
Actor Entropy Scale & $\eta$ & $3 \times 10^{-4}$ \\
Learning Rate & $\alpha$ & $4 \times 10^{-5}$ \\
Adam Betas & $\beta_{1}$, $\beta_{2}$ & 0.9, 0.999 \\
Adam Epsilon & $\epsilon$ & $1 \times 10^{-20}$ \\
Gradient Clipping & AGC & 0.3 \\
\midrule
Per method loss weight & & \\
Dreamerv3 reconstruction loss & --- & 1.0 \\
NE-Dreamer next-embedding loss & --- & 1.0 \\
R2-Dreamer barlow-twins loss & --- & 0.05 \\
DreamerPro SwAV loss & --- & 1.0 \\
\midrule
NE-Dreamer transformer configuration & & \\
hidden dim & --- & 256 \\
num layers & --- & 2 \\
num heads & --- & 4 \\
\midrule
Additional hyperparameters &  --- & \\
BT redundancy $\lambda$ & --- & $5 \times 10^{-4}$ \\
\bottomrule
\end{tabular}
\label{table:hyperparams}
\end{table*}

\section{DMC detailed results}
Figure~\ref{app:dmc} the individual learning curves for all 20 tasks in the DMC benchmark.
\label{app:dmc}
\begin{figure*}[h]
  \centering
  \includegraphics[width=\linewidth]{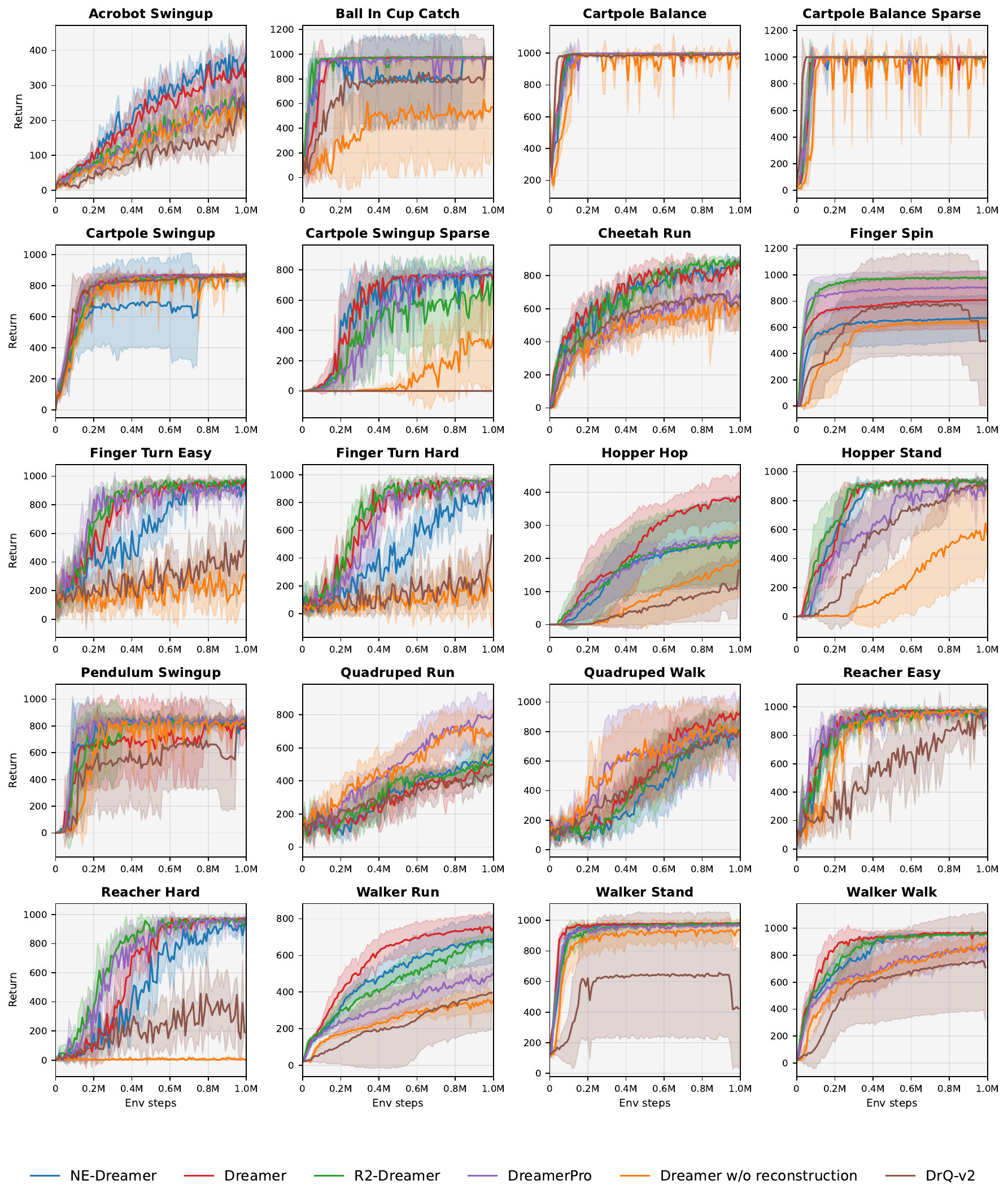}
  \caption{Per-task learning curves for all 20 DMC tasks}
  \label{fig:dmc_per_task_comparison}
\end{figure*}

\end{document}